# A Novel Approach to Distributed Multi-Class SVM

Aruna Govada[1], Shree Ranjani[2], Aditi Viswanathan[3], S.K.Sahay[4]

garuna@goa.bits-pilani.ac.in ,shreeranjaniss@gmail.com,aditivin@gmail.com,sashay@ goa.bits-pilani.ac.in

Department of CS&IS, BITS Pilani K. K. Birla Goa Campus,

Zuarinagar, Goa, PIN-403726, India.

## Abstract

With data sizes constantly expanding, and with classical machine learning algorithms that analyze such data requiring larger and larger amounts of computation time and storage space, the need to distribute computation and memory requirements among several computers has become apparent. Although substantial work has been done in developing distributed binary SVM algorithms and multi-class SVM algorithms individually, the field of multi-class distributed SVMs remains largely unexplored. This research proposes a novel algorithm that implements the Support Vector Machine over a multi-class dataset and is efficient in a distributed environment (here, Hadoop). The idea is to divide the dataset into half recursively and thus compute the optimal Support Vector Machine for this half during the training phase, much like a divide and conquer approach. While testing, this structure has been effectively exploited to significantly reduce the prediction time. Our algorithm has shown better computation time during the prediction phase than the traditional sequential SVM methods (One vs. One, One vs. Rest) and out-performs them as the size of the dataset grows. This approach also classifies the data with higher accuracy than the traditional multi-class algorithms.

## Keywords

Distributed algorithm, Support Vector Machine, Machine learning, Mapreduce, Multi class

## 1. Introduction and Related Work

In the machine learning world, SVMs offer one of the most accurate results. SVMs are accurate because of their high generalization property to classify unknown examples. Yet SVM algorithms have been largely restricted to simple 2-class (binary) classification problems. However, numerous practical applications involve multi-class classifications - like identifying the galaxy that a star belongs to, remote sensing applications, etc. Some of the most used multi-class SVM approaches include One vs One, One vs Rest, DAG and Error correcting codes (all of which have their own drawbacks and are not as efficient as binary SVM algorithms).

In One vs. Rest classification, the n-class problem is converted into n 2-class sub problems with one positive class and (n-1) negative classes. In One vs. Rest classification, the n-class problem is converted into n(n-1)/2 two-class problems. Krebel [1] showed that by this formulation, unclassifiable regions reduce, but still they remain. To solve the problem of unclassifiable regions, Taylor et al. [2] proposed decision-tree based pairwise classification Graph. Pontil et al. [3] proposed to use rules of a tennis tournament to solve unclassified regions. Kiksirikul et al. [4] proposed the same method and called it Adaptive Directed Acyclic Graph. A comparison of these approaches [5] suggest the usefulness of One vs One in terms of accuracy and computation and this is why we have chosen to compare our approach with this.

For both binary SVMs as well as multi class SVMs, in the recent years, handling large datasets has become an arduous task. Data Scientists are overwhelmed with the amount of data and the need for excessive data pre-processing that this explosion has caused. Given that data handling has become tough, data mining – the process of discovering new patterns from large data datasets – is a herculean task. This has given rise to scientists developing distributed parallel algorithms to meet the scalability and performance requirements for big data. Computation time and computation complexity (which involves solving the quadratic optimization problem) has been a limiting factor for SVMs especially for large data sets. To overcome this, many parallel and distributed SVMs were proposed. Initially most of the parallel SVM was based on MPI programming model. Moving from the MPI programming model based parallel SVM, parallelization has been achieved through the MapReduce Framework now. Fox [6] developed parallel SVM based on iterative MapReduce model Twister. A parallelization scheme was proposed where the kernel matrix is approximated by a block-diagonal approach [7]. Further improvements to parallel SVM implementations like Cascade SVMs [8] have been proposed which heavily reduce the communication overhead among the computers. In this method, dataset is split into parts in feature space. Non-support vectors of each sub dataset are filtered and only support vectors are transmitted. Collobert et al. [9] proposed a new parallel SVM training and classification algorithm that each subset of a dataset is trained with SVM and then the classifiers are combined into a final single classifier function. Lu et al. [10] proposed a connected network based distributed support vector machine algorithm. In this method, the dataset is split into roughly equal part for each computer in a network then, support vectors are exchanged among these computers. Sun et al. proposed a novel method for parallelized SVM based on MapReduce technique. This method is based on the cascade SVM model. Their approach is based on iterative MapReduce model Twister which is different from our implementation which is a recursive MapReduce algorithm. Ferhat et al. [11] proposed a novel MapReduce based binary SVM training method in which the whole training dataset is distributed over data nodes of cloud computing system using Hadoop streaming and MRjob python library. Despite such extensive work on multi class SVMs as well as distributed binary SVMs, the arena of multi class distributed SVMs has remained largely unexplored. In this paper, we propose a novel algorithm for distributed multi class SVMs and have compared our results with the most popular multi class SVM approaches (One vs. One and One vs. Rest)

## 2. Proposed Framework

The proposed algorithm is based on binary tree kind of structure created during the training phase. Our algorithm aims to reduce the total number of SVMs required to classify a data point, thus enabling better efficiency during run-time of the model that was built out of our algorithm. While One vs. One, One vs. Rest and DAGSVM classify using $\frac{n(n-1)}{2}$, $n$, and $\frac{n(n-1)}{2}$ SVMs respectively, we use $log_2(n+1)$ SVMs to classify the data point at run-time. One possible structure that can be obtained in depicted in Figure II. It is critical to choose the most appropriate combination to obtain the most optimal case while testing for a new sample data point. For this, we have separated the training stage into 2 significant phases where the first stage (Training) is devoted to compute all possible support vectors and the second stage (Cross Validation) evaluates all of them and returns the best division.

### 2.1 Training

Given N classes, we partition the entire dataset into 2 halves each containing [n/2] classes using support vectors. This is done neglecting the differences among the classes on one side. Without loss of generality, one half has been assigned as positive class and the other negative class. To generate support vectors, Atbrox's [12] method for parallel machine learning has been used which gave us the mapper and reducer implementation for binary classification. Atbrox's method implements incremental SVM algorithm for binary classification as described below:

The SVM classifier solves the following problem of finding w,y i.e. the coefficients of the support vector formulated as

$$(w, y) = (I/\mu + E^T E)^{-1} E^T D e$$

Where I – identity matrix
  $\mu$ - parameter >0
  E = [A –e]
  D – Diagonal matrix with plus ones or minus ones

To classify a test sample with feature vector x, following equation is used.

$$sgn\ (x^T w - y) = \begin{cases} 1, & x \in A+ \\ -1, & x \in A- \end{cases}$$

Where A+ and A- denote the positive and negative classes respectively.
Mappers and reducers have been used to parallelize the calculation of $E^T E$ and $E^T D e$ and Figure I. depicts a brief outline of the algorithm which explains the function of each mapper and reducer used in this approach.
As at any point, binary classification is performed where each class represents many, mappers and reducers from Atbrox have been modified to suit our purpose. A single run of the training stage is as follows:
- Divide the dataset into 2 regions using ($^nC_{n/2}$) / 2 planes where 'n' stands for the number of classes in the dataset. This figure is arrived based on the intuition that a plane divides the data points into roughly half the number of classes on each side. i.e choose n/2 out of n and and to avoid repeated counting , the number of possible combinations was divided by 2.
- For all possible combinations support vectors are formed.

## 2.2 Cross Validation

This stage of training primarily involves identifying the best plane from the possible options obtained from previous stage.
A single run of the second stage is as follows.
- For each of the partitions thus obtained, accuracy with which each plane divides is calculated using the classification accuracy metric ((true positives + true negatives)/total samples).
- Mappers split the task of obtaining the confusion matrix (The matrix which contains true positives, true negatives, false positives, false negatives). Reducers assimilate the values in the confusion matrix from each node and compute the classification accuracy metric. This metric is used to identify the best split and store the 2 separated lists of classes for further computation.
- At the end of this second stage, we obtain a set of positive and negative classes along with their corresponding accuracy calculation.

Both the stages are repeated until the number of classes in the positive and negative become one, which effectively means that the dataset has been successfully divided into all N classes.

## 2.3 Testing

The classification of a test sample starts at the root of the tree. At each node of the binary tree a decision is being made about the assignment of the input pattern into one of the two possible groups obtained after the training phase. Each of these groups may contain multiple classes. This is repeated recursively downward the tree until the sample reaches a leaf node that represents the class it has been assigned to (Figure II). Any test sample will go through a maximum of $log_2 N$ SVMs during the test phase.

Compute the $(^nC_{n/2})/2$ binary combinations of the set of classes.
**Eg:** For a 3 class problem (with classes - 1,2 and 3), the combinations will be:

| | | (+ve classes | -ve classes) |
|---|---|---|---|
| 1,2 | 3 | | |
| 1,3 | 2 | | |
| 2,3 | 1 | | |

…………..(i)

**STEP 1: Training** of SVMs with the Training Dataset.
**Output** – 1 SVM is learned for each of the combinations listed in (i) …………..(ii)
Refer to (A) below for details

**STEP 2: Cross Validation** of SVMs with the Cross Validation Dataset.
**Output** – 1 most accurate SVM is chosen from those learned in (ii) ………….(iii)
Refer to (B) below for details

No of classes ==1? → STOP

The SVM computed from (iii) is added to the final tree structure.
$(^nC_{n/2})/2$ combinations are computed again for each of the set of +ve and –ve classes recursively.

**Eg:** Suppose combination -
2,3     1
gave the most accurate SVM.
In the next step, combinations would be:
2     3

**(A)**
D - matrix of training classes(1.0(+veclass), -1.0 (-ve class))
A - matrix with feature vectors,
e - vector filled with ones,
E = [A -e]
mu = scalar constant # used to tune classifier
**MAPPER**
**Input:** Rows of data from the Training Dataset
**Computation:** Matrices E.T*E and E.T*D*e
**Output:** Base 64 encoded (E.T*E, E.T*D*e)

**REDUCER**
**Input:** Output of mapper (key – index of combinations of classes for which SVM is being computed – from cases.txt; value – base 64 encoded E.T*E, E.T*D*e)
**Computation:** (Large margin with linear kernel): (omega, gamma) = inverse(I/mu + sum(E.T*E)) * sum(E.T*D*e)
**Output:** Coefficients (thetas) for SVM computation for each line in cases.txt

**(B)**
(omega, gamma) = (I/mu + E.T*E).I*(E.T*D*e)
x –incoming feature vector
**MAPPER**
**Input:** Rows of data from the Cross Validation Dataset
**Computation:** x.T*omega - gamma
**Output:** Number of true positives, true negatives, false positives and false negatives for each case in cases.txt

**REDUCER**
**Input:** Output of mapper (key – index of combinations of classes for which SVM is being computed – from cases.txt; value – TPs, FPs, TNs, FNs)
**Computation:** Accuracy, Precision, Recall, F1 measure for each case in cases.txt. Selection of best SVM based on this.
**Output:** Coefficients of SVM with the best accuracy/F1 measure

Figure I. A brief outline of the algorithm

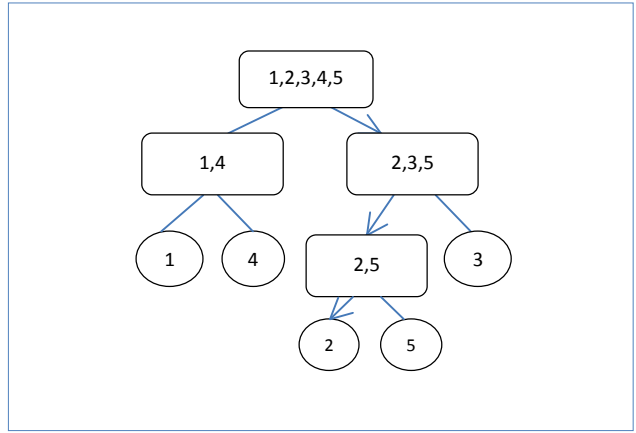

**Figure II**: This structure is created during the training phase. A test sample belonging to class 2 will follow the path depicted by the arrows

## 3. Experiments and Results

For all of these experiments, we have used a 3 node cluster to measure the metrics of our approach, and Python's Scikit-learn library for One vs. One and One vs. Rest). Datasets used for experimentation are described below and the sources for those are indicated in references.

### 3.1 Datasets used

| Dataset name | SDSS[15] | Iris[16] | Mfeat[17] |
|---|---|---|---|
| # Training samples | 40000 | 150 | 1500 |
| # Testing samples | 10000 | 50 | 500 |
| # Features | 6 | 3 | 6 (mor) 47 (zer) 64 (kar) |
| # Classes | 3 | 4 | 10 |

**Figure III**: Datasets used

### 3.2 Accuracy (Figure IV.)

We have measured the accuracy of the algorithm using the following formula on the testing samples

I. $Accuracy = \frac{TPs + TNs}{Total\ Samples}$

Our approach gives better accuracies in all the datasets except the SDSS dataset. SDSS is a skewed dataset, so accuracy is not the best performance metric to use during cross-validation. We will have to use metrics other than the accuracy (such as precision, recall and F1 measure) to select the best SVM here

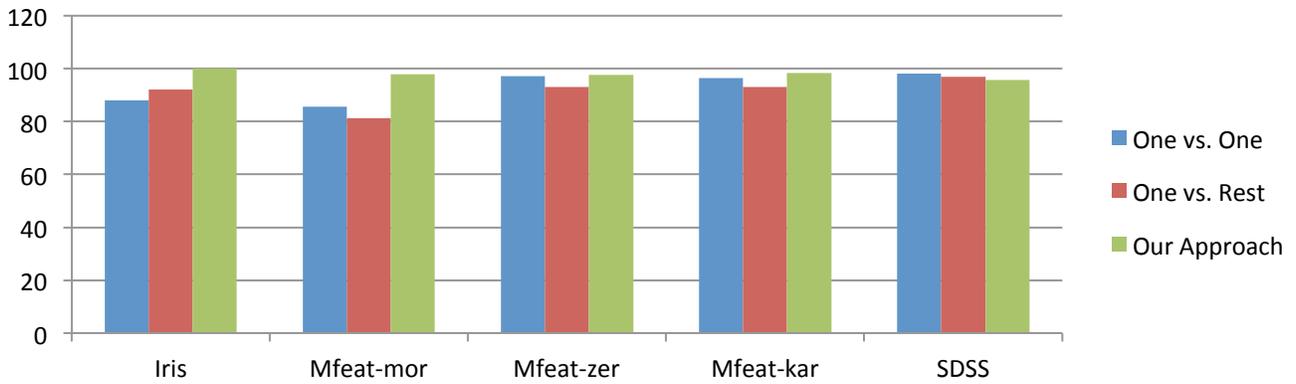

### 3.3 Training Time (Figure V.)

While the single-machine implementations are more efficient for the smaller datasets, in the SDSS dataset we see that our training time is comparable to the single-machine implementations due to the large data size of SDSS. We can thus show that distribution of the computation gets more beneficial as the data size increases.

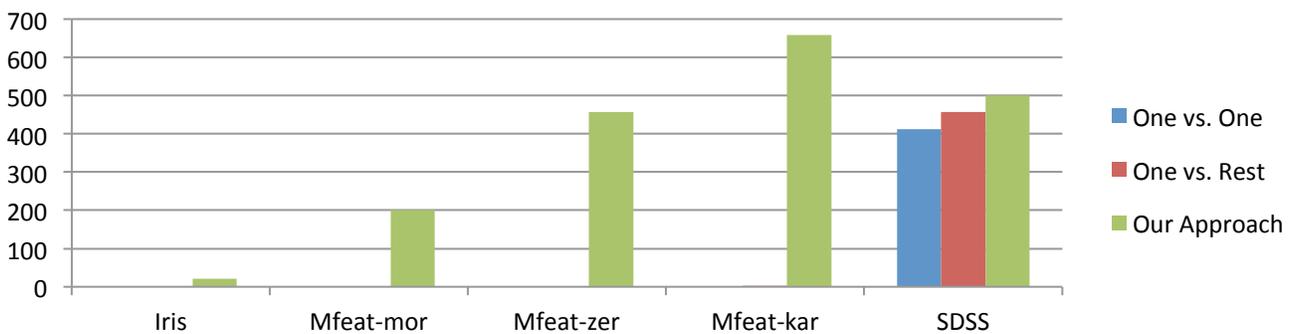

### 3.4 Testing Time (Figure VI.)

We show a significant reduction (53.7%) in testing time for the SDSS dataset, a result of the distributed approach working hand-in-hand with the decision tree based algorithm.

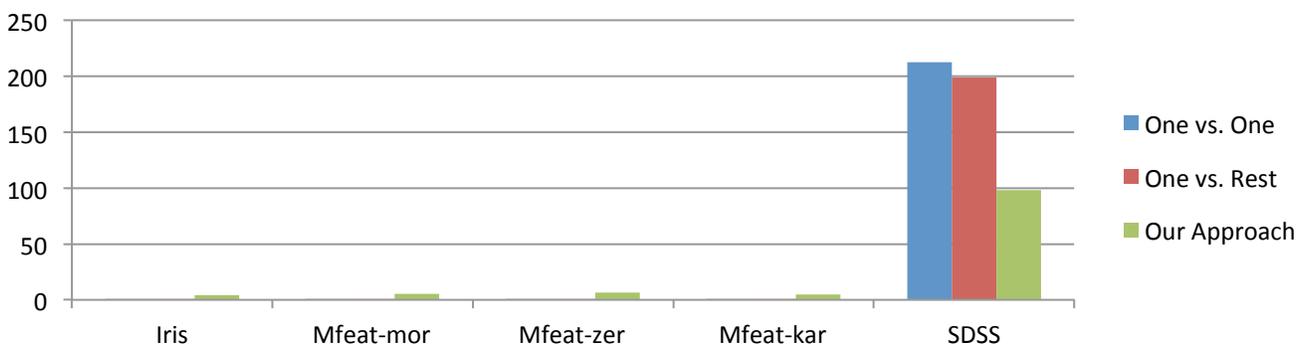

In this research, we have proposed a novel distributed multi class SVM algorithm in which instead of extending binary SVMs or all-together methods, the idea is to divide the dataset into half at any point of time and obtain the visual distribution during the training phase. While testing, this structure has been effectively exploited and hence saving huge amount of testing time. This approach has been found to excel as data size increases which caters to our needs of handling big data.

In the future, we hope to enhance this algorithm by doing the following:

- Implementing a distributed Gaussian Kernel (we are currently using a linear kernel)
- Optimizing the algorithm for skewed datasets by using performance metrics such as the F1 measure (instead of Accuracy that we use currently)
- Running the algorithm with data sizes of about 20-30 GB with very large clusters (which we haven't been able to do so far for lack of resources)
- Comparing this algorithm with other multi-class Machine Learning techniques (non-SVM)